\documentclass[12pt]{article}
\usepackage[T2A]{fontenc}
\usepackage[cp1251]{inputenc}
\usepackage{amsmath,amsfonts,amssymb}
\usepackage[mathscr]{eucal}
\usepackage{graphicx}
\usepackage{subcaption}
\usepackage{algorithm}
\usepackage[noend]{algpseudocode}

\usepackage{hyperref}
\usepackage[capitalize,noabbrev]{cleveref}

\newcounter{zacountsec}
\newcommand{\eh}{\hfill}\newlength{\sperr}

\newcommand{\Section}[1]{{\stepcounter{zacountsec}\vspace{3mm}%
\hspace*{18mm}\normalsize\bf\arabic{zacountsec}.\parbox[t]{150mm}{#1}}}

{\end{minipage}}
\newenvironment{thm}[2]{\begin{sloppypar}%
{#1 #2.}\em{}}%
{\end{sloppypar}}

\newcommand{\proof}{\hspace*{9mm}{\settowidth{\sperr}{\rm  Proof
}%
\parbox[t]{1.3\sperr}{\rm  P\eh r\eh o\eh o\eh f} }}

\newcommand{\R}{{\rm  I}\!{\rm  R}}
\newcounter{zalit}

{\end{list}\end{small}}
\newlength{\addro}\newlength{\addrt}
{\end{minipage}}

\def\R{{\mathbb R}}

\def\L2{{L_2(\Gamma)}}
\def\e2{{e_2(g,\,\Gamma)}}

\newcommand{\ls}{\left(}
\newcommand{\rs}{\right)}

\newcommand{\la}{\left\langle}
\newcommand{\ra}{\right\rangle}

\DeclareMathOperator{\diag}{diag}
\DeclareMathOperator{\diagonal}{diagonal}

\makeatletter
\renewcommand*{\@biblabel}[1]{\hfill#1.}
\makeatother

\begin{document}
\vspace*{0mm}

\begin{center}
\large\bf STOCHASTIC GRADIENT DESCENT WITH PRECONDITIONED POLYAK STEP-SIZE \\ 
\vspace*{1mm}
\end{center} 

\vspace*{1mm}
\begin{center}
{
Farshed Abdukhakimov $^{a,1}$, Chulu Xiang $^{a,2}$, \\ Dmitry Kamzolov $^{a,3}$, and Martin Tak\'a\v{c} $^{a,4}$}\\
\vspace*{3mm}
{\it $^{a}$ 
Mohamed bin Zayed University of Artificial Intelligence, Abu Dhabi, UAE \\
$^{1}$e--mail: farshed888@gmail.com \qquad 
$^{2}$e--mail: chulu.xiang@mbzuai.ac.ae \\ 
$^{3}$e--mail: kamzolov.opt@gmail.com \qquad 
$^{4}$e--mail: takac.mt@gmail.com \\ 

} 
\medskip
\end{center}

\noindent
{\small {\bf Abstract} -- 
Stochastic Gradient Descent (SGD) is one of the many iterative optimization methods that are widely used in solving machine learning problems. These methods display valuable properties and attract researchers and industrial machine learning engineers with their simplicity. However, one of the weaknesses of this type of methods is the necessity to tune learning rate (step-size) for every loss function and dataset combination to solve an optimization problem and get an efficient performance in a given time budget. Stochastic Gradient Descent with Polyak Step-size (SPS) is a method that offers an update rule that alleviates the need of fine-tuning the learning rate of an optimizer. In this paper, we propose an extension of SPS that employs preconditioning techniques, such as Hutchinson's method, Adam, and AdaGrad, to improve its performance on badly scaled and/or ill-conditioned datasets. 
}

\vspace*{2mm}

\vspace*{3mm}

\begin{center}
{\large 1. INTRODUCTION}
\end{center}

In this paper, we consider an Empirical Risk Minimization (ERM) problem known as finite-sum optimization
\begin{equation}
w^{\ast} \in \mathop{\arg\min}\limits_{w \in \mathbb{R}^d    } \{f(w):=\tfrac{1}{n}\sum_{i=1}^n f_i(w)\},
\label{eq:1}
\end{equation}
where $w \in \mathbb{R}^d$ is the weight parameter and each $f_i:\mathbb{R}^d\rightarrow \mathbb{R}$ is a smooth and twice differentiable objective function. The loss function $f_i(w)=$ computes the difference between the prediction of a model with weights parameters $w$ and a target value $y$. 
The  objective is to then minimize the average loss $f(w)=\frac{1}{n}\sum_{i=1}^n f_i(w)$ over a given data with $n$ elements denoted as $\{(x_i,y_i)\}_{i=1}^n$, where $x_i$ is an input data point and $y_i$ is its corresponding target label. Due to the complicated nature of this minimization problem the closed form solution can rarely be found, hence forcing one to resort to alternative optimization algorithms that can minimize it using iterative or other methods. One of the iterative methods designed for this minimization is \textit{Stochastic Gradient Descent} (SGD), which updates the weight parameters as follows
\begin{equation}
    w_{t+1} = w_t - \gamma_t \nabla f_i(w_t),
    \label{eq:SGD1}
\end{equation}
where $\nabla f_i(w_t)$ is the mini-batched stochastic gradient and $\gamma_t\in \R$ is the step-size (or learning rate) of the update. Using mini-batches of a large dataset significantly reduces convergence time to an optimal point $w^{\ast}$. 
Extensive research has been conducted on stochastic first-order methods begging from the pioneering papers by Robbins and Monro \cite{robbins1951stochastic}, Polyak \cite{polyak1990new}, Polyak and Juditsky \cite{polyak1992acceleration}, Nemirovski et al.\cite{nemirovski2009robust} and accelerated version by Lan \cite{lan2012optimal}. Let us stress, that every loss function and dataset combination requires special tuning of a step-size $\gamma_t$ to find a minimum, which turns $\gamma_t$ into a hyperparameter.  This issue of fine-tuning the hyperparameter $\gamma_t$ was one of the motivations behind methods with an adaptive learning rate, where $\gamma_t$ is replaced by an expression which adaptively changes during optimization process. In recent times, there has been a widespread of adaptive techniques that dynamically modify the step size \cite{adagrad,kingma2014adam,li2019convergence,reddi2019convergence,ward2020adagrad,loshchilov2018decoupled,shi2023aisarah,sadiev2022stochastic}, especially for enhancing the training of deep neural networks.

Another line of adaptive stochastic methods, Stochastic Gradient Descent with Polyak Step-size (\textit{SPS}) was inspired by step-size policy proposed by Boris T. Polyak in \cite{polyak1969minimization, polyak1987introduction} for subgradient methods. Later, in \cite{berrada2020training,loizou2021stochastic} stochastic version of Polyak Step-size was proposed with different extensions \cite{orvieto2022dynamics, gower2022cutting,li2022sp2,schaipp2023stochastic, schaipp2023momo, garrigos2023function,jiang2023adaptive}. We will discuss it in more details in the next section. 
\begin{equation}
    \gamma_t = \frac{f_i(w_t) - f_i^{\ast}}{\|\nabla f_i(w_t)\|^2}
    \label{eq:sps_f}
\end{equation}

One of the main topics discussed in this paper is deriving methods that are designed to overcome badly scaled/ill-conditioned datasets and, \textit{preconditioning} -- is a technique that can be employed for this problem. Even though ideal preconditioning is practically infeasible to achieve, we can still utilize various methods: \textit{Hutchinson's method} and techniques used in other adaptive methods such as Adam \cite{kingma2014adam} and AdaGrad \cite{adagrad}.

\vspace{2mm}
{\bf \large Notation.}
We endow the primal space $w\in \mathbf{E}$ and dual space $g \in \mathbf{E}^{\ast} $ by the conjugate norms: $\|w\|$ and $\|g\|_{\ast}$. As a special case, for a positive definite matrix $B \in \mathbb{R}^{d \times d}$ we define conjugate Euclidean norms: $\|w\|_B=\la Bw,w\ra^{1/2} $ and $\|g\|_{B^{-1}} = \la g, B^{-1}g\ra^{1/2}$, where $\nabla f(w) \in \mathbf{E}^{\ast} $ and $\nabla^2 f(w)h \in \mathbf{E}^{\ast}$. The operator $\odot$ is defined as a component-wise product between two vectors, also known as the Hadamard product. We use $\diag(v)$ as a diagonal matrix of a given vector $v$ and a vector $\diagonal(\mathbf{H}) \in \mathbb{R}^{d}$ as the diagonal of a matrix $\mathbf{H}\in\mathbb{R}^{d\times d}.$

\begin{center}
{\large 2. RELATED WORK}
\end{center}

Let us define a generic update rule for algorithms that we will be analyzing:
\begin{equation}
    w_{t+1} = w_t - \gamma_t M_t m_t,
    \label{eq:update_precond}
\end{equation}
where $\gamma_t$ is the learning rate, $M_t$ is the preconditioning matrix derived with certain rule, and $m_t$ is either $g_t$ (true gradient or gradient approximation) or the first moment of the gradient with momentum parameter $\beta_1$. To interpret this update, one can imagine a search direction $m_t$ to be scaled and rotated by the preconditioning matrix $M_t$ and the step-size $\gamma_t$. Some of the adaptive first-order optimization methods display their update rule in a slightly different manner:
\begin{equation}
    w_{t+1} = w_t - \gamma_t m_t / v_t,
    \label{eq:moments_update_precond}
\end{equation}
where $m_t$ and $v_t$ are called first and second moment terms respectively. These two updates incorporate the same concept of preconditioning the search direction and can be used interchangeably throughout this paper for convenience. 

Classical optimization methods can also be displayed in this fashion. For instance, for SGD the preconditioning matrix $M_t=I$, while $m_t=g_t$ and $\gamma_t$ is set to a constant. We would like to note that $\gamma_t$ in SGD is one of the particularly important and cumbersome hyperparameters since it needs to be tuned according to ones specific data and loss functions. To address this, methods with adaptive learning rate were introduced and some of them utilize the preconditioning matrix which is based on the local curvature of the loss function. 

It is important to overview the fundamental part of methods with {\bf Polyak step-size}. To derive the classical deterministic Polyak step-size let us consider a convex function $f(w_t)$ and the distance of the $w_{t+1}$ to the optimal solution $w^{\ast}$ which is upper-bounded as follows:
\begin{align*}
    \| w_{t+1} - w^{\ast} \|^2 \leq Q(\gamma), \quad \text{where} \quad Q(\gamma) = \| w_t - w^{\ast} \|^2 - 2\gamma [ f(w_t) - f^* ] + \gamma_t^2 \| g_t \|_{\ast}^2.
\end{align*}
Here $g_t$ denotes the subgradient of a function $f(w_t)$ and $f^*$ the optimal function value. Minimizing the upper-bound $Q(\gamma)$ we can obtain the Polyak step-size and express it in terms of the update rule \eqref{eq:update_precond}:
\begin{equation}
    \gamma_t = \mathop{\arg\min}\limits_{\gamma} [Q(\gamma)] = \frac{f(w_t) - f^*}{\| g_t \|_{\ast}^2}, \quad M_t=I \quad \text{and} \quad m_t=g_t.
    \label{eq:polyak}
\end{equation}
Detailed proof can be found in \cite{boyd2003subgradient}. One can notice that the step-size \eqref{eq:polyak} can only be used when the optimal solution $f^*$ is known. Even though some problems might have $f^* = 0$ as the optimal solution, the deterministic nature of this method makes it not applicable. To avoid this limitation of the deterministic Polyak step-size, \textit{Stochastic Gradient Descent with Polyak step-size} (SPS) \cite{loizou2021stochastic} was proposed alongside with a more practical version SPS\textsubscript{max} which restricts the $\gamma_t$ from being too large: 
\begin{equation}
    \gamma_t^{\text{SPS}} = \frac{f_i(w_t) - f_i^*}{ \| \nabla f_i(w_t) \|_{\ast}^2}, \quad \text{and} \quad \gamma_t^{\text{SPS}\textsubscript{max}} = \min \Big\{ \frac{f_i(w_t) - f_i^*}{ \| \nabla f_i(w_t) \|_{\ast}^2}, \gamma_b \Big\}
    \label{eq:sps}
\end{equation}
SPS still requires the knowledge of $f_i^*$, but in optimization of standard unregularized loss functions, such as squared loss for regression and logistic loss for classification, the optimal solution $f_i^*$ is equal to $0$. In terms of the update rule \eqref{eq:update_precond} we can display SPS as:
\begin{equation}
    \gamma_t = \gamma_t^{\text{SPS}}, \quad M_t=I \quad \text{and} \quad m_t=\nabla f_i(w_t).
    \label{eq:sps_update}
\end{equation}

There is another way of deriving the SPS method. If we assume that the interpolation condition holds, then we can solve \eqref{eq:1} by sampling $i\in \{1,\ldots,n\}$ i.i.d at each iteration $t$ and then solving the nonlinear equation
\begin{equation}
    w_{t+1} = \mathop{\arg\min}\limits_{w \in \mathbb{R}^d} \| w - w^t \|^2 \quad \text{s.t.} \quad f_i(w) = 0.
    \label{eq:3}
\end{equation}
While the above projection might have a closed form solution for some simple loss functions, for most nonlinear models like Deep Neural Networks (DNNs) there is no closed-form solution of \eqref{eq:3}.
So instead of solving it exactly, we can linearize the $f_i(w)$  around the current iterate $w^t$ to obtain
\begin{gather}
    w_{t+1} = \mathop{\arg\min}\limits_{w \in \mathbb{R}^d} \| w - w^t \|^2 \quad \text{s.t.} \quad f_i(w^t) + \langle \nabla f_i(w^t), w - w^t \rangle  = 0.
\end{gather}
Update rule \eqref{eq:sps_update} serves as an exact closed-form solution for this problem.

Outside of the interpolation regime there might not exist a solution for \eqref{eq:3}. 
    So instead of trying to set all loss functions to zero, we can try making them all small by minimizing a \emph{slack} variable as follows
    \begin{gather}
        \textstyle\mathop{\arg\min}\limits_{w \in \mathbb{R}^d, s \geq 0} \quad s \nonumber\\
        \text{s.t.} \quad f_i(w) \leq s, \quad \text{for} \quad i = 1, \dots, n,
        \label{eq:slack}
\\
        \textstyle\mathop{\arg\min}\limits_{w \in \mathbb{R}^d, s \geq 0} \quad s^2 \nonumber\\
        \text{s.t.} \quad f_i(w) \leq s, \quad \text{for} \quad i = 1, \dots, n,
        \label{eq:slack_sq}
    \end{gather}
    which are referred to as \textit{L1} and \textit{L2} slack minimization~\cite{gower2022cutting}, respectively. One can note that the goal of this method is to force $s$ to be as small as possible which  allows to solve problems where the interpolation assumption does not hold or the model is under-parameterized.

\begin{center}
{\large 3. CONTRIBUTION}
\end{center}
Here we combine preconditioning and variants of slack regularized SPS methods. We then demonstrate that these new preconditioned methods perform well on badly scaled and ill-conditioned data.

\begin{itemize}
    \item \textbf{Updated SPS. \;} We extend the SPS methods and present 3 updated algorithms \textit{PSPS}, \textit{PSPSL1} and \textit{PSPSL2} which use \textit{Hutchinson's} method of Hessian estimation, Adam and AdaGrad to precondition search directions and include the scaling of Polyak step-size with a weighted Euclidean norm. Closed-form updates to our methods are described later.
    
    \item \textbf{PyTorch Implementation.\;} We develop practical variants of our methods as \textit{PyTorch} optimizers and make the code publicly available at our GitHub repository\footnote{\url{https://github.com/fxrshed/ScaledSPS}.}.
 
    \item \textbf{Empirical Results.\;} Several experiments are conducted in 2 different settings to compare our results to SGD, Adam, AdaGrad and to variants of SPS that are not applying any preconditioning techniques. We demonstrate  the proposed algorithms exhibit noticeable improvements on badly scaled data. 
\end{itemize}

\begin{center}
{\large 4. PRECONDITIONING}
\end{center}

Data can be badly scaled and/or ill-conditioned and preconditioning is one way to improve the convergence speed of algorithms. Algorithms that take advantage of preconditioning have a generic update rule as following
\begin{equation}
    w_{t+1} = w_t - \gamma_t M_t \nabla f_i(w_t),
\end{equation}
 where $M_t \in \mathbb{R}^{d \times d} $ is 
 an invertible positive definite matrix.
A textbook example of a method that utilizes this technique is Newton's method where $M_t = \nabla^2 F(w_t)$ and $\gamma_t=1$. More recent and practical methods include AdaHessian, Adagrad and OASIS 
\cite{yao2021adahessian,adagrad,jahani2022doubly}. These methods incorporate curvature of the loss function via adaptive estimates of the Hessian. 

\begin{center}
{\large 4.1 HUTCHINSON'S METHOD}
\end{center}
\label{subsection:hutch}

Hutchinson's method \cite{hutchinson1989stochastic} is used to estimate  the diagonal of the Hessian matrix.
To compute this estimate, the Hutchinson method uses only a few Hessian-vector products, which in turn can be computed efficiently using backpropagation~\cite{Christianson:1992}.
Indeed, the product of a Hessian matrix $\textbf{H}=\nabla^2 f(w)$ and a fixed vector $z$ can be computed through a directional derivative of the gradient. To understand how this method is used for preconditioning first we show that the computation costs of Hessian-vector product is close to $2$ computations of gradients, i.e.,
\begin{equation}
\nabla^2 f(w) z = \nabla ( z^T \nabla f(w)).
\label{eq:hvp}
\end{equation} 
Then, we can compute the Hessian diagonal using Hutchinson's method:
\begin{equation}\label{eq:hutchinson}
\diag(\textbf{H})=\mathbb{E}[z\odot (\textbf{H}z)],
\end{equation}
where $z$ is a random vector with Rademacher distribution\footnote{$z_i \in \{-1, +1\}$ with equal probability.} or a normal distribution and $\textbf{H}z$ is computed by the Hessian matrix oracle given in \eqref{eq:hvp}. 
It can be proved that the expectation of $z\odot (\textbf{H}z)$ is the Hessian diagonal (see~\cite{BEKAS20071214}). 
Using this identity, we estimate the Hessian diagonal from a given $D_0$  by sampling a vector $z$ at each iteration, 
and iteratively updating our estimate using a weighted average as follows
\begin{equation}
    D_t = \beta D_{t-1} + (1 - \beta) \diag(z \odot \textbf{H}z),
\end{equation}
where $\beta \in (0, 1)$ is a momentum parameter and
\begin{equation}
    D_0 = \frac{1}{m} \sum_{i=1}^{m} \diag(z_i \odot \textbf{H}_i z_i),
\end{equation}
where $\textbf{H}_i$ denotes a Hessian at the initial point $w_0$ of a randomly sampled batch. Finally, to ensure that $D_t$ remains Positive Definite, despite possible  non-convexity of the loss functions, we use truncation and keep only absolute values of elements as follows
$      ( \hat{D}_t  )_{i, i} = \max \{ \alpha, | D_t |_{i, i} \}$.

\begin{algorithm}[H]
\begin{algorithmic}[1]
\State {\bf Inputs:} $\beta \in (0, 1)$, $\alpha > 0$
\State {\bf Initialize:} 
$D_0 = \frac{1}{m} \sum_{i=1}^{m} \diag(z_i \odot \textbf{H}_i z_i)$
\For{$t =1,\ldots, T-1$}
\State Sample $z$ from Rademacher/Normal distribution
\State  $D_t = \beta D_{t-1} + (1 - \beta) \diag(z \odot \textbf{H}z)$
\State $( \hat{D}_t  )_{i, i} = \max \{ \alpha, | D_t |_{i, i} \}$
\EndFor  
\State {\bf Output:} $D$
\end{algorithmic}
\caption{Hessian Diagonal Estimation using Hutchinson's Method}
\label{alg:hutch}
\end{algorithm}
\clearpage

\begin{center}
    {\large 4.2 AdaGrad}
\end{center}

AdaGrad is a stochastic optimization method that approximates the Hessian of the optimized function in order to adapt the learning rate depending on the curvature information. The key idea involves using the cumulative squared gradient information to scale the learning rates. Hence, in terms of the update \eqref{eq:moments_update_precond}, the update rule for AdaGrad can be is given by:

\begin{align}
\label{eq:adagrad_update}
m_t = \mathbf{g_t} , \quad \text{and} \quad v_t = \sqrt{ \sum_{i=1}^t \mathbf{g_i} \mathbf{g_i} }.
\end{align}
Accumulation of all previous gradients in the preconditioner $v_t$ leads to decay in the learning rate $\gamma_t$ which increases performance for sparse settings (non-frequent features) at the cost of degrading in case of dense settings. 

\begin{center}
    {\large 4.3 Adam}
\end{center}

Introduced in \cite{kingma2014adam}, Adam is designed to overcome the limitations of other popular optimization algorithms, such as AdaGrad \cite{adagrad} and RMSProp \cite{tieleman2012lecture}, by incorporating both adaptive learning rates and momentum-based updates. The update rule of Adam involves the computation of the moving average of both the first and second moments of the gradients. The first moment is the mean of the gradients, and the second moment is the uncentered variance of the gradients. The update rule for Adam can be expressed in terms of the update \eqref{eq:moments_update_precond} as follows:
\begin{align}
\label{eq:adam_update}
m_t     &= \frac{(1 - \beta_1)\sum_{i=1}^t \beta_1^{t-i} \mathbf{g_i} }{1 - \beta_1^t}, \\ 
v_t     &= \sqrt{ \frac{(1 - \beta_2)\sum_{i=1}^t \beta_2^{t-i} \mathbf{g_i}\mathbf{g_i} }{1 - \beta_2^t}},
\end{align}
where $0 < \beta_1, \beta_2 < 1$ are two hyperparameters referred to as first and second moment coefficients.  The biased estimates are corrected by dividing them by the bias correction terms, which are powers of the decay rates $\beta_1$ and $\beta_2$, respectively.

\begin{center}
    {\large 5. PRECONDITIONED STOCHASTIC POLYAK STEP-SIZE}
\end{center}

In this section we propose new methods that are inspired by the previously described methods such as SPS. First of all, in order to describe them we note that we consider the projection and constraint 
\begin{equation}
     w_{t+1} = \mathop{\arg\min}\limits_{w \in \mathbb{R}^d} \| w - w_t \|^2 \quad \text{s.t.} \quad f_i(w) = 0.
     \label{eq:proj}
\end{equation}
Note that the constraint $f_i(w) = 0$ is defined as the \textit{interpolation condition} as defined as follows

\vspace*{2mm}
{\bf Definition 1.}
{\it 
We assume that the \textit{interpolation} condition holds for a set of functions
  $\{f_i(w)\}_{i=1}^n$ over a given dataset $\{(x_i,y_i)\}_{i=1}^n$ with a non-negative loss functions, $f_i(w) \geq 0$, when
    \begin{equation}
    \exists w^{\ast} \in \mathbb{R}^d \quad \text{s.t.} \quad f_i(w^{\ast}) = 0, \quad \forall i \in \{1,2,\dots,n\}.      
    \end{equation}
}

One of the presented techniques used in our work is utilizing preconditioning in order to get a better convergence rate in case of badly scaled data. To develop this, we change the norm in the projection \eqref{eq:proj} to a weighted norm based on the preconditioning matrix $B_t \succ 0$. Another important idea is linear approximation of the interpolation condition $f_i(w) = 0$. According to Taylor expansion of a function $f_i(w)$ the linear (first-order) approximation is given by $f_i(w) \approx f_i(w_t) + \langle \nabla f_i(w_t), w-w_t\rangle$. We use this approximation to relax the interpolation condition which does not allow the closed-form solution for most of the nonlinear models. Another way to solve this problem is to introduce a \textit{slack} variable (described later). 

\vspace{2mm}
{\bf Preconditioned SPS.} We consider a differentiable convex function $f_i$ and a linearization of the interpolation condition. To derive a preconditioned update rule we use a weighted norm in the projection which we refer to as PSPS (\textbf{P}reconditioned \textbf{S}tochastic Gradient Descent with \textbf{P}olyak \textbf{s}tep-size). In this paper we consider 3 preconditioning techniques discussed previously, namely Hutchinson's method and preconditioning of AdaGrad and Adam optimizers. 

\vspace{2mm}
{\bf Lemma 1. (PSPS)}
{\it 
\label{lemma:psps}
Let $B_t \succ 0$ for any $t\geq 0$.
Then the iterative update of the following problem
\begin{equation}
w_{t+1} = \mathop{{\arg \min}}\limits_{w \in \mathbb{R}^d} \  \tfrac{1}{2} \|w - w_t\|_{B_t}^2  \quad \mbox{s.t.} \quad 
  f_i(w_t) + \langle \nabla f_i(w_t), w-w_t\rangle = 0 \nonumber
\end{equation}
is given by
\begin{equation}
    w_{t+1} = w_t - \tfrac{f_i(w_t)}{{||\nabla f_i(w_t)||}_{B_t^{-1}}^2} B_t^{-1} \nabla f_i(w_t).
    \label{eq:psps}
\end{equation}
}
Note that this update rule can be reformulated in terms of the update rule \eqref{eq:update_precond}, where 
\begin{equation}
    \gamma_t = \tfrac{f_i(w_t)}{{||\nabla f_i(w_t)||}_{B_t^{-1}}^2}, \quad M_t = B_t^{-1} \quad \text{and} \quad m_t = \nabla f_i(w_t).
\end{equation}

Similarly, we can apply preconditioning into the slack based methods and derive two methods which we refer to as \textit{PSPSL1} and \textit{PSPSL2}.

\vspace{2mm}
{\bf Lemma 2. (PSPSL1)}
{\it
Let ${B_t} \succ 0$ for any $t\geq 0$ and $\mu, \lambda > 0$. Then the closed-form update for the following problem 
\begin{align}
    w_{t+1}, s_{t+1} = \mathop{{\arg \min}}\limits_{w\in\mathbb{R}^d,s\geq
    0}\tfrac{1}{2}\|w-w_t\|_{B_t}^2+\mu(s-s_t)^2+\lambda s \nonumber \\
    \text{s.t.} \quad f_i(w_t)+\left\langle \nabla f_i(w_t),w-w_t\right\rangle\leq s,
    \label{eq:pspsl1}
\end{align}
is given by 
\begin{align}
     \gamma _t^{L1} &= \tfrac{(f_i(w_t)-s_t+\tfrac{\lambda}{2\mu})_+}{\tfrac{1}{2\mu}+\|\nabla f_i(w_t)\|_{B_t^{-1}}^2},\nonumber \quad \gamma _t =\min \{\gamma _t^{L1} ,\tfrac{f_i(w_t)}{\|\nabla f_i(w_t)\|_{B_t^{-1}}^2} \},\nonumber \\
    w_{t+1} &= w_t-\gamma_t B_t^{-1} \nabla f_i(w_t), \quad s_{t+1} = (s_t-\tfrac{1}{2\mu}(\lambda+\gamma _t^{L1}))_+.
\label{eq:spsl1solu}
\end{align}

}

\vspace{2mm}
{\bf Lemma 3. (PSPSL2)}
{\it
Let $B_t \succ 0$ for any $t\geq0$ and $\mu, \lambda > 0$. Then the closed form update for the following problem
\begin{align}
\label{eq:pspsl2}
    w_{t+1},s_{t+1}=\mathop{{\arg \min}}\limits_{w\in\mathbb{R}^d,s\in \mathbb{R}} \|w-w_t\|_{B_t}^2+
    \mu(s-s_t)^2+\lambda s^2 \nonumber \\
    \text{s.t.} \quad f_i(w_t)+\left\langle \nabla f_i(w_t),w-w_t\right\rangle\leq s, 
\end{align}
is given by
\begin{align}
    w_{t+1}&=w_t-\tfrac{(f_i(w_t)-\mu\hat{\lambda }s_t)_+}{\hat{\lambda }+\|\nabla f_i(w_t)\|_{B_t^{-1}}^2}B_t^{-1}\nabla f_i(w_t), \\
    s_{t+1} &=\hat{\lambda } \ls \mu s_t+\tfrac{(f_i(w_t)-\mu \hat{\lambda }s_t)_+}{\hat{\lambda }+\|\nabla f_i(w_t)\|_{B_t^{-1}}^2} \rs,
\end{align}
\label{eq:spsl2solu}
where $\hat{\lambda}=\tfrac{1}{\mu+\lambda}$.
}
Here slack parameter $\lambda$ forces $s$ to be closer to 0 while $\mu$ does not allow $s_{t+1}$ to be far from $s_t$.

\begin{center}
{\large 6. NUMERICAL EXPERIMENTS}
\end{center}

In this section we present the experiments conducted using our proposed methods and some of the most popular optimizers: SGD, Adam and AdaGrad. The choice of these methods is justified be the fact that all of these methods, except SGD, employ adaptive learning rate. In our experiments each of these methods are presented with different step-sizes to show the difference in convergence. 

We used LIBSVM\footnote{\url{https://www.csie.ntu.edu.tw/~cjlin/libsvmtools/datasets/}} datasets, namely \textit{mushrooms} and \textit{colon-cancer}, to illustrate the performance of proposed methods minimizing Logistic Regression and Non-Linear Least Squares loss functions on binary classification problems. Furthermore, every experiment is additionally conducted on \textit{badly scaled} version of the same datasets, where the columns are multiplied by a vector  $e = \{\exp(x_i)\}_{i=1}^d$ where $x_i$ is generated from a uniform distribution on the interval $[-k, k]$. In the following illustrations the term $k$ refers to this scaling factor, where $k=0$ is original data. 

During training all the proposed methods we applied slack parameters $\lambda=0.01$ and $\mu=0.1$. For Hutchinson's method we applied $\alpha=1 \times 10^{-4}$ and $\beta=0.999$. Hyperparameters (except of step-size) for other methods (SGD, Adam, etc.) were kept as default values. All experiments were run with $5$ different seeds using \textit{PyTorch 1.11.0}.

{\bf Loss Functions.}
Let ${\{(x_i,y_i)\}}_{i=1}^n$ be our dataset.  \textit{Logistic regression} is defined as
$    f_{LogReg}(w) = \frac{1}{n}\sum_{i=1}^n \log(1 + \exp(-y_i x_i^T w))$, where $x_i \in \mathbb{R}^d$ and $y_i \in \{-1, +1\}$ and \textit{non-linear least squares} is given by
$    f_{NLLSQ}(w) = \frac{1}{n}\sum_{i=1}^n ( y_i - 1 / (1 + \exp( -x_i^T w ) ) )^2,
$
where $y_i \in \{0, 1\}$.

\begin{figure} 
    \centering
    \begin{subfigure}{0.49\textwidth}
    \includegraphics[width=\textwidth]{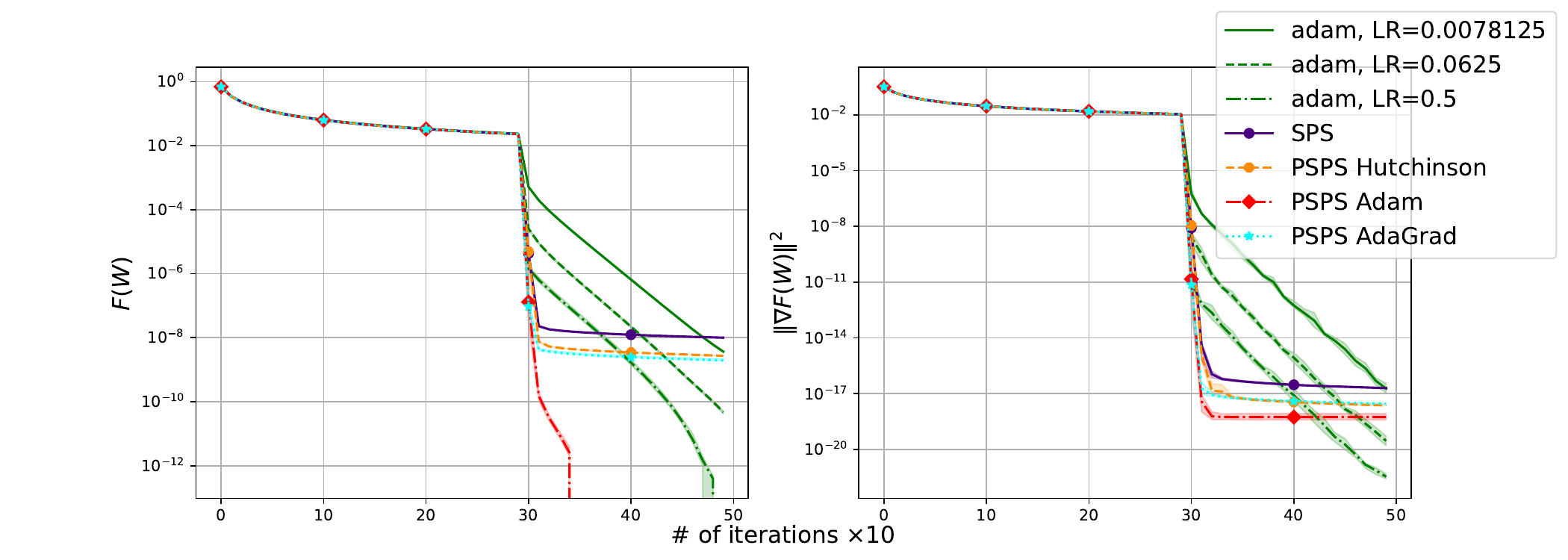}
    \caption{$k = 0$}
    \label{fig:adam-psps-mushrooms-logreg-0}
    \end{subfigure}
    \hfill
    \begin{subfigure}{0.49\textwidth}
    \includegraphics[width=\textwidth]{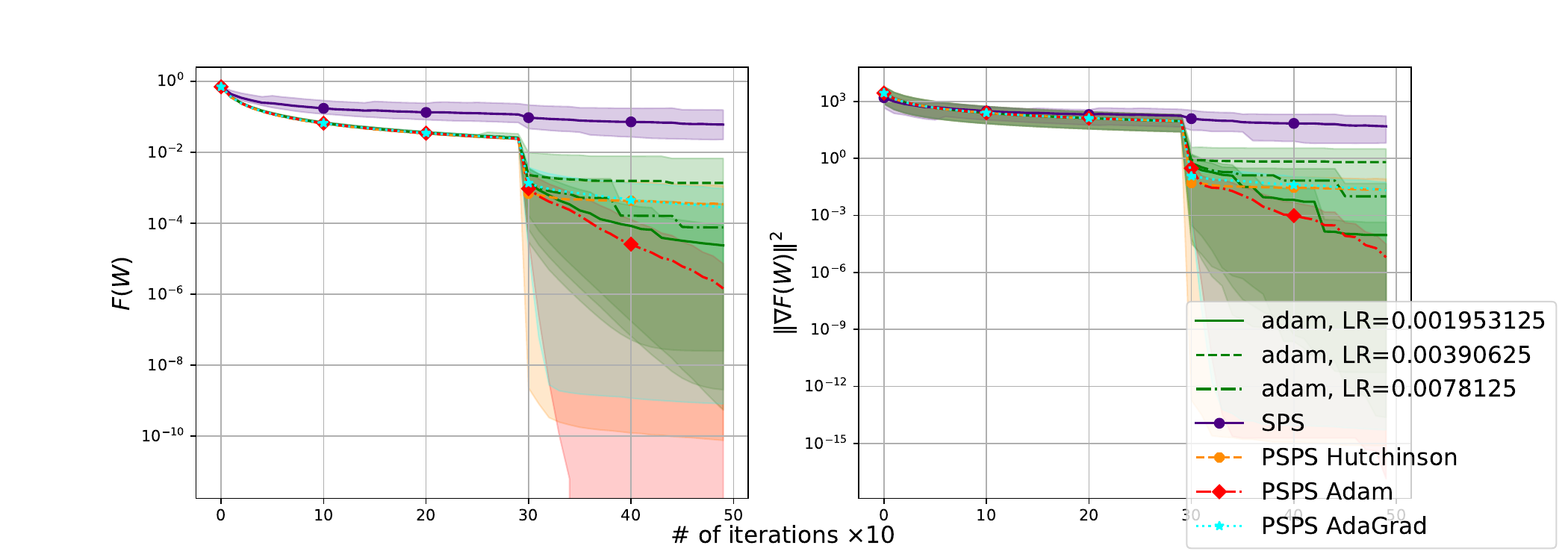}
    \caption{$k = 6$}
    \label{fig:adam-psps-mushrooms-logreg-6}
    \end{subfigure}
    \caption{Adam vs PSPS methods with different preconditioning methods on \textit{mushrooms} dataset with \textit{logistic regression} loss function.}
    \label{fig:adam-psps-mushrooms-logreg}
\end{figure}

\begin{figure}
    \centering
    \begin{subfigure}{0.49\textwidth}
    \includegraphics[width=\textwidth]{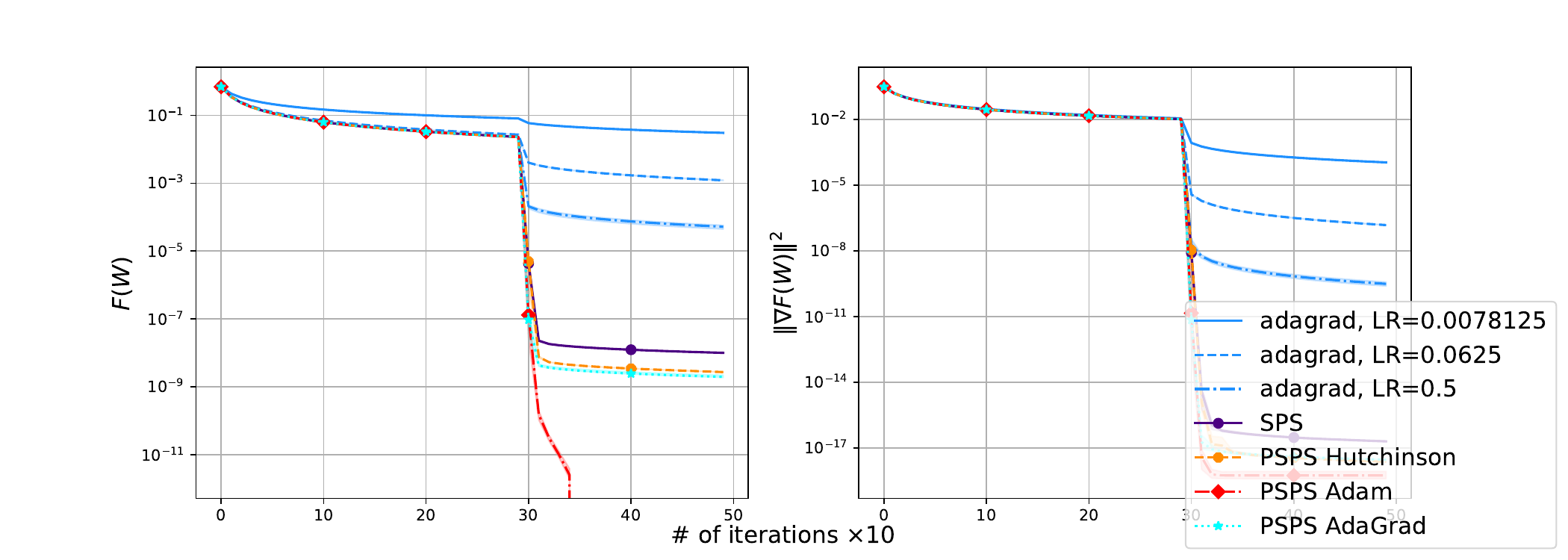}
    \caption{$k = 0$}
    \label{fig:adagrad-psps-mushrooms-logreg-0}
    \end{subfigure}
    \hfill
    \begin{subfigure}{0.49\textwidth}
    \includegraphics[width=\textwidth]{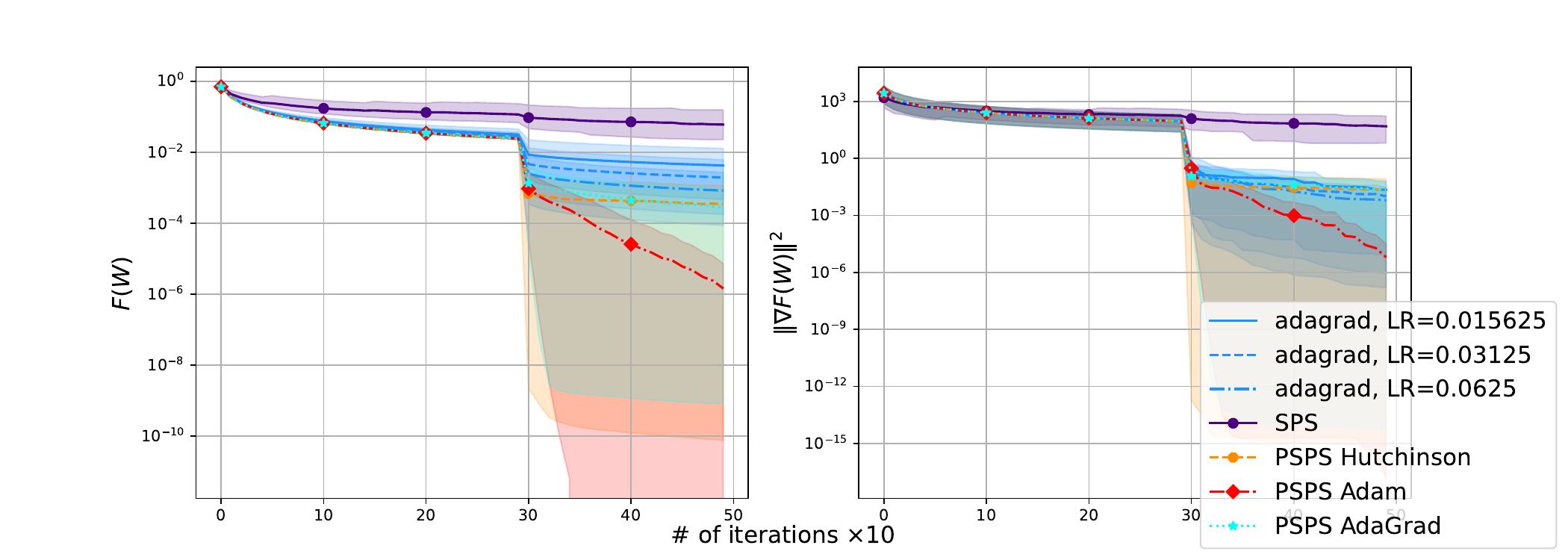}
    \caption{$k = 6$}
    \label{fig:adagrad-psps-mushrooms-logreg-6}
    \end{subfigure}
    \caption{AdaGrad vs PSPS methods with different preconditioning methods on \textit{mushrooms} dataset with \textit{logistic regression} loss function.}
    \label{fig:adagrad-psps-mushrooms-logreg}
\end{figure}

\begin{figure}
    \centering
    \begin{subfigure}{0.49\textwidth}
    \includegraphics[width=\textwidth]{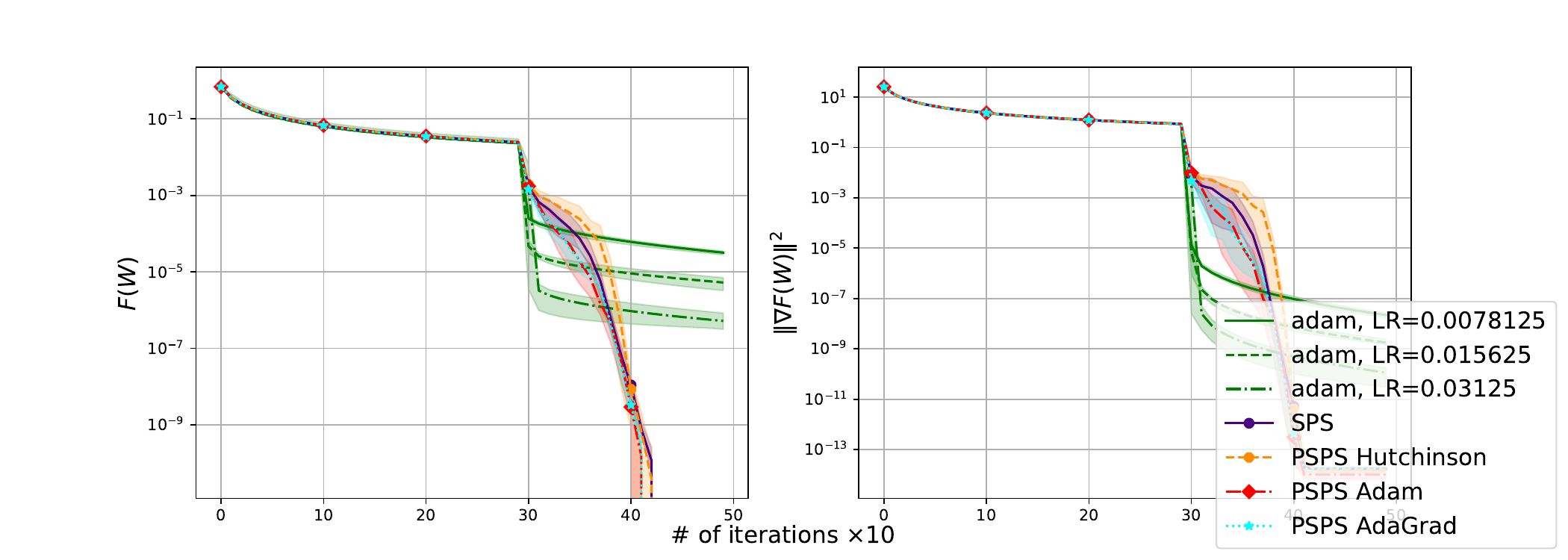}
    \caption{$k = 0$}
    \label{fig:adam-psps-colon-cancer-logreg-0}
    \end{subfigure}
    \hfill
    \begin{subfigure}{0.49\textwidth}
    \includegraphics[width=\textwidth]{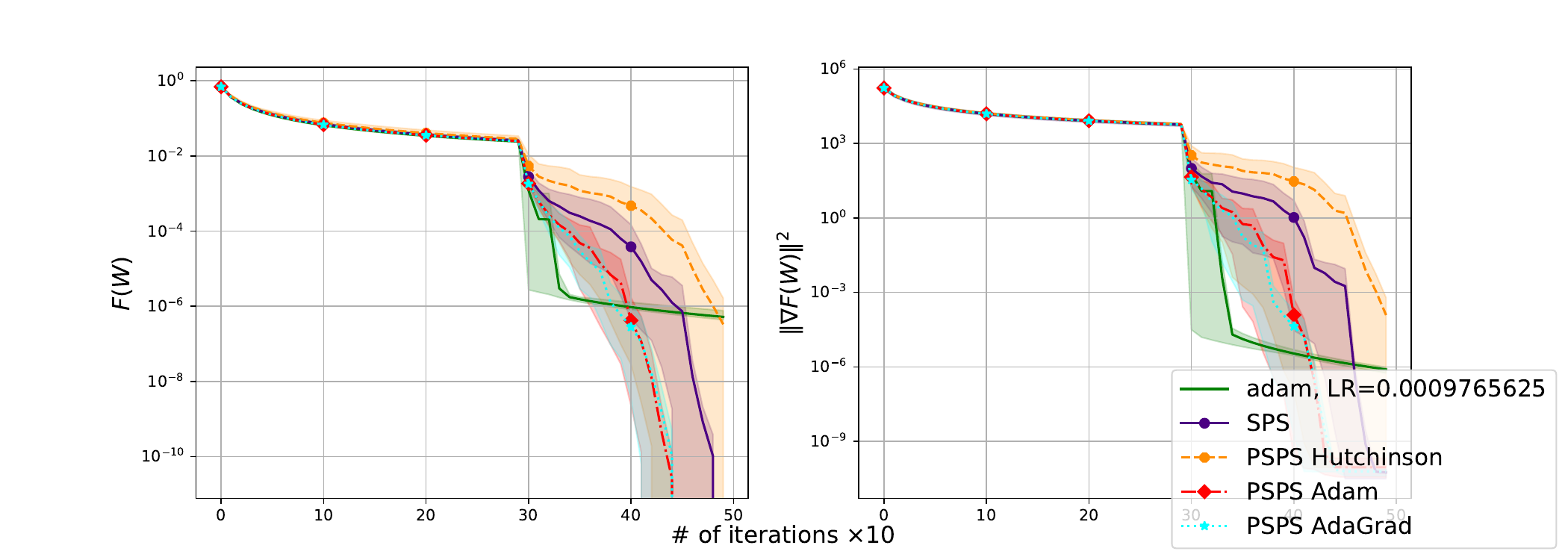}
    \caption{$k = 6$}
    \label{fig:adam-psps-colon-cancer-logreg-6}
    \end{subfigure}
    \caption{Adam vs PSPS methods with different preconditioning methods on \textit{colon-cancer} dataset with \textit{logistic regression} loss function.}
    \label{fig:adam-psps-colon-cancer-logreg}
\end{figure}

In Figure~\ref{fig:adam-psps-mushrooms-logreg} we compare convergence rates of SPS with and without preconditioning against Adam. We observe that in case of badly scaled version of the dataset we are required to fine-tune the learning rate of the Adam optimizer in order to not diverge since keeping the same learning rates in both cases resulted in divergence in case of $k = 6$. Also, we can see how different preconditioning techniques outperform SPS without any preconditioning in both original dataset and badly scaled. Not requiring any learning rate manual fine-tuning is one of the advantages of preconditioned SPS methods. Similar results can be observed in Figure~\ref{fig:adagrad-psps-mushrooms-logreg} and in Figure~\ref{fig:adam-psps-colon-cancer-logreg} on \textit{colon-cancer} dataset. In Figure~\ref{fig:adam-psps-colon-cancer-logreg-6} we can see that scaling the dataset results in Adam optimizer's learning rate fine-tuned to be much smaller as scaling factor increases in order to not diverge.

We also compare our results to original \textit{SPS}, \textit{SPSL1}, \textit{SPSL2}, \textit{SGD}, and \textit{Adam}.

\begin{figure}
    \centering
    \begin{subfigure}{1.\textwidth}
    \includegraphics[width=\textwidth]{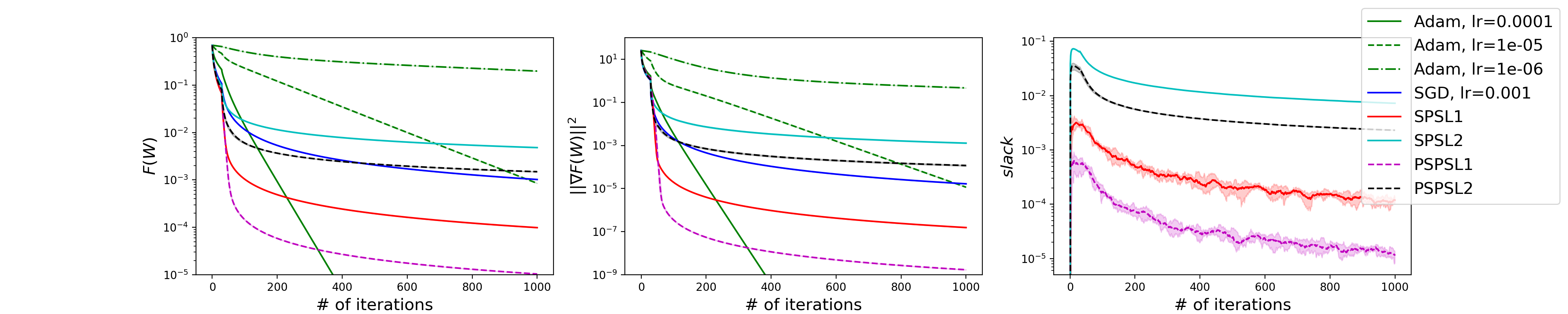}
    \caption{$k = 0$}
    \label{fig:psps-slack-colon-cancer-logreg-0}
    \end{subfigure}
    \hfill
    \begin{subfigure}{1.\textwidth}
    \includegraphics[width=\textwidth]{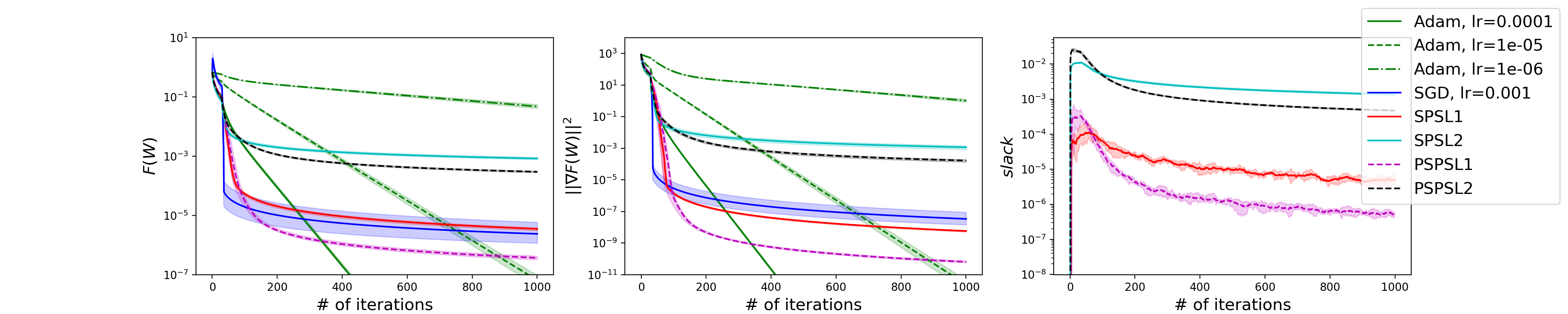}
    \caption{$k = 3$}
    \label{fig:psps-slack-colon-cancer-logreg-6}
    \end{subfigure}
    \caption{Performance comparison of PSPSL1 and PSPSL2 to non-scaled versions of SPS, SGD and Adam on original and badly scaled versions of \textit{colon-cancer} dataset. All methods are trained on Logistic Regression.}
    \label{fig:psps-slack-colon-cancer-logreg}
\end{figure}

\begin{figure}
    \centering
    \begin{subfigure}{1.\textwidth}
    \includegraphics[width=\textwidth]{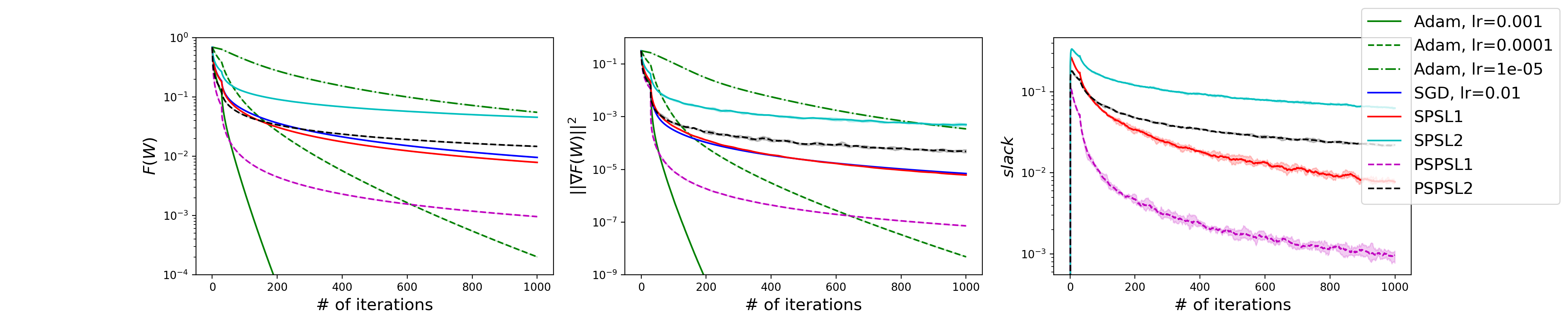}
    \caption{$k = 0$}
    \label{fig:psps-slack-mushrooms-logreg-0}
    \end{subfigure}
    \hfill
    \begin{subfigure}{1.\textwidth}
    \includegraphics[width=\textwidth]{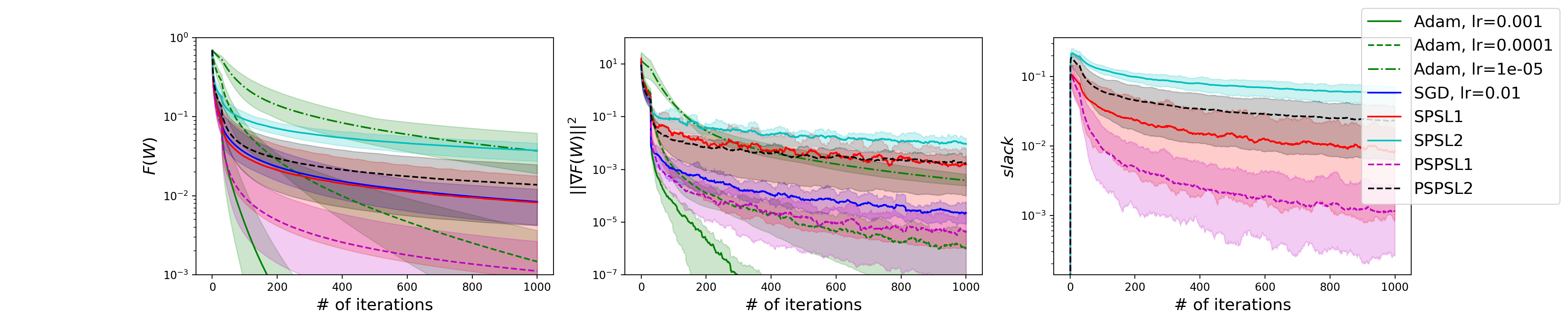}
    \caption{$k = 3$}
    \label{fig:psps-slack-mushrooms-logreg-6}
    \end{subfigure}
    \caption{Performance comparison of PSPSL1 and PSPSL2 to non-scaled versions of SPS, SGD and Adam on original and badly scaled versions of \textit{mushrooms} dataset. All methods are trained on Logistic Regression.}
    \label{fig:psps-slack-mushrooms-logreg}
\end{figure}

\clearpage

\begin{center}
{\large 6. CONCLUSION AND FUTURE WORK}
\end{center}

In this paper we studied the effect of preconditioning on the family of SPS(Stochastic Gradient Descent with Polyak Step-size) methods. We showed modified update rules \textit{PSPS, PSPSL1, PSPSL2} in \ref{eq:pspsl1} and \ref{eq:pspsl2}. In our solution a new parameter $\mu$ is introduced which helps to control the step direction of slack $s$. Experiments were conducted in both convex and non-convex settings with 2 different datasets. \\
\textbf{Future work.\;}This paper lacks theoretical analysis of our proposed methods which can be done as a follow up research work. On top of that, it is highly encouraged to extend experiments to a realm of Deep Neural Networks.


\bibliography{ref,kamzolov_full}

\newpage
\clearpage

\appendix

\section{Appendix}

\subsection{Proof of Lemma 1 (PSPS)}
Let 
\begin{align*}
      w^{\ast}=\arg\min_{w\in\mathbb{R} ^d}\frac{1}{2} \|w - w_t\|_{B_t}^2 \nonumber \\
      \textit{s.t.} \quad f_i(w_t) + \langle \nabla f_i(w_t), w-w_t\rangle = 0 \nonumber.
\end{align*}

To simplify notation, we denote that 

\begin{equation}
  a=\frac{1}{\sqrt{2}}(w-w_t), \nonumber
\end{equation}

\noindent Then,

\begin{align}
\min &\frac{1}{2}||a||_{B_t}^2 \notag \\
s.t. \quad f_i(w_t) &+  \la\nabla f_i(w_t), a\ra=0. \notag
\end{align}

We introduce $\lambda$, and let $ L=\frac{1}{2}||a||_{B_t}^2+\lambda(f_i(w_t)+\la\nabla f_i(w_t), a\ra) $, taking the partial derivation with respect to $a$ and $\lambda$ ,
\begin{align}
    \frac{\partial L}{\partial a}&=B_t a+\lambda \nabla f_i(w_t)=0 \nonumber
    \\
    \frac{\partial L}{\partial \lambda}&=f_i(w_t)+ \la\nabla f_i(w_t), a\ra=0.
    \nonumber
\end{align}

To solve these equations, we get
\begin{align}
a &=-\lambda B_t^{-1} \nabla f_i(w_t) \nonumber\\
\lambda & = \frac{f_i(w_t)}{\|\nabla f_i(w_t)\|_{B_t^{-1}}^2} .\nonumber
\end{align}

Finally,
\begin{equation}
  \hat w=w_t-\frac{f_i(w_t)}{{\|\nabla f_i(w_t)\|}_{B_t^{-1}}^2} B_t^{-1} \nabla f_i(w_t). \nonumber 
\end{equation}

\subsection{Proof of Lemma 2 (PSPSL1)}
We can rewrite the slack part of the objective function in \eqref{eq:pspsl1} as
\begin{equation}
   \lambda s+\mu(s-s_t)^2=\frac{1}{2}\cdot 2\mu \ls s-s_t+\frac{\lambda}{2\mu}\rs ^2 +
   constants\; w.r.t\; w\; and\; s.
\end{equation}
Dropping constants independent of $s$ and $w$ and let $s^0=s_t-\frac{\lambda}{2\mu}$  we have that () is equivalent to solving
\begin{equation}
  w_{t+1},s_{t+1}={\arg\min}_{w\in\mathbb{R}^d,s\geq 0}
\|w-w_t\|_{B_t}^2+2\mu(s-s^0) ^2  \nonumber
\end{equation}
\begin{equation}
   f_i(w_t)+\left\langle \nabla f_i(w_t),w-w_t\right\rangle - (s-s^0)   \leq s^0 \nonumber
\end{equation}
\begin{equation}
    s\geq0.
\end{equation}
(1) If $s^0\geq f_i(w_t)$ holds then the solution is simply $(w_{t+1},s_{t+1})=(w_t,s^0)$.  \\ (2) If $s^0\leq f_i(w_t)$. And at least one of the inequality constraints must be active at the optimal point as the problem is an L2 projection onto the intersection of two halfspace.\\
(\romannumeral1) If the constraints 
$
   f_i(w_t)+\left\langle \nabla f_i(w_t),w-w_t\right\rangle
= s 
$ is active, and let $w-w_t=\alpha$, $s-s^0=\beta$, then our problem reduces to 
\begin{equation}
  \alpha_{t+1},\beta_{t+1}={\arg\min}_{w\in\mathbb{R}^d,s\geq 0}
\|\alpha\|_{B_t}^2+2\mu\beta^2  \nonumber
\end{equation}
\begin{equation}
   f_i(w_t)+\left\langle \nabla f_i(w_t),\alpha\right\rangle-\beta = s^0\nonumber
\end{equation}
\begin{equation}
-s^0-\beta \leq 0
\end{equation}
Let $L=\|\alpha\|_{B_t}^2+2\mu\beta^2+\theta(f_i(w_t)+\la\nabla f_i(w_t), \alpha\ra-\beta-s^0)+\gamma(-s^0-\beta)$, and take the derivative with respect to $\alpha$, $\beta$. And get KKT conditions:
\begin{equation}
    \begin{cases}
       \frac{\partial L}{\partial \alpha}=2B_t\alpha
    +\theta\nabla f_i(w_t)=0\\
   \frac{\partial L}{\partial \beta}=4\mu\beta-\theta-\gamma=0\\
   \gamma \geq 0\\
   f_i(w_t)+  \la \nabla f_i(w_t),\alpha\ra -\beta - s^0=0\\
   \gamma(-s^0-\beta)=0,
   \end{cases}
\end{equation}
which is equivalent to
\begin{equation}
    \begin{cases}
       \alpha=-\frac{\theta B_t^{-1}\nabla f_i(w_t)}{2}\\
        \beta=\frac{\theta+\gamma}{4\mu}\\
       \theta=\frac{4\mu(f_i(w_t)-\frac{\gamma}{4\mu}-s^0)}{1+2\mu {\|\nabla f_i(w_t)\|}_{B_t^{-1}}^2}\\
       \gamma\geq\frac{-2s^0(2\mu{\|\nabla f_i(w_t)\|}_{B_t^{-1}}^2)-2f_i(w_t)}{\|\nabla f_i(w_t)\|_{B_t^{-1}}^2}\\
       \gamma\geq 0.
    \end{cases}
\end{equation}

So when condition $2\mu\|\nabla f_i(w_t)\|_{B_t^{-1}}^2  s^0+f_i(w_t)\geq0$ holds, then the solution is given by
\begin{equation}
    \begin{cases}
       \beta=\frac{\theta}{4\mu}\\
       \alpha=-\frac{\theta B_t^{-1}\nabla f_i(w_t)}{2}\\
       \theta=\frac{4\mu(f_i(w_t)-s^0)}{1+2\mu {\|\nabla f_i(w_t)\|}_{B_t^{-1}}^2}.
    \end{cases}
\end{equation}
which is equivalent to
\begin{equation}
\begin{cases}
    w-w_t=-\frac{(f_i(w_t)-s^0)}{\frac{1}{2\mu}+{\|\nabla f_i(w_t)\|}_{B_t^{-1}}^2}B_t^{-1}\nabla f_i(w_t) \\
    s-s^0= \frac{(f_i(w_t)-s^0)}{1+2\mu{\|\nabla f_i(w_t)\|}_{B_t^{-1}}^2}.
\end{cases}
\end{equation}

If not, we have solution as following:
\begin{equation}
    \begin{cases}
       \beta=-s^0\\
       f_i(w_t)+  \la \nabla f_i(w_t),\alpha\ra =0.
    \end{cases}
\end{equation}
This problem can be solved similarly as proof of Lemma 1, and its solution is given by
\begin{equation}
\begin{cases}
   w-w_t=-\frac{f_i(w_t)}{{\|\nabla f_i(w_t)\|}_{B_t^{-1}}^2}B_t^{-1}\nabla f_i(w_t)\\
   s-s^0=-s^0.
\end{cases}
\end{equation}

(\romannumeral2)
If the constraints $s_{t+1}=0$ is active then our problem reduces to
\begin{equation}
    \min_{w\in \mathbb{R}^d} \|w-w_t\|_{B_t}^2 \nonumber
\end{equation}
\begin{equation}
    f_i(w_t)+\left\langle \nabla f_i(w_t),w-w_t\right\rangle\leq 0
\end{equation}

which is a projection onto a halfspace, and its solution is given by
\begin{equation}
    w-w_t=-\frac{f_i(w_t)}
{{\|\nabla f_i(w_t)\|}_{B_t^{-1}}^2}B_t^{-1}\nabla f_i(w_t)
\end{equation}
To sum up all these above cases can be written as solution which is given by Lemma 2 \eqref{eq:pspsl1}.

{\bf Lemma 2.1 }
\label{lemma:lemma5}
Let $ \delta> 0$,$  c\in \mathbb{R} $ and $w,w^0,a\in\mathbb{R}^d$. The closed-form solution to 
\begin{equation}
    w',s'={\arg\min}_{w\in\mathbb{R}^d,s\in\mathbb{R}^b}\|w-w^0\|_{B_t}^2+\delta(s-s^0)^2 \nonumber
\end{equation}
s.t.
\begin{equation}
    a^T(w-w^0)+c\leq s,
\end{equation}
is given by
\begin{equation}
    w'=w^0-\delta\frac{(c-s^0)_+}{1+\delta\|a\|_{B_t^{-1}}^2}B_t^{-1}a,
\nonumber
\end{equation}
\begin{equation}
    s'=s^0+\frac{(c-s^0)_+}{1+\delta\|a\|_{B_t^{-1}}^2}.
\label{eq:lemma4}
\end{equation}

{\bf Proof }
Let  $\alpha=w-w^0,\beta=s-s^0$, then our question becomes 
\begin{equation}
\alpha,\beta=\arg\min_{\alpha\in\mathbb{R}^d,\beta\in\mathbb{R}^b}\|\alpha\|_{B_t}^2+\delta\beta^2 \nonumber
\end{equation}
s.t.
\begin{equation}
    a^T\alpha-\beta+c-s^0\leq 0.
\end{equation}
(1)If $w=w^0$ and $s=s^0$ satisfies in the linear inequality constraint, that is if $c\leq s^0$, in which case the solution is simply $w'=w^0$ and $s'=s^0$.\\
(2) But if $c\geq s^0$, $(w^0,s^0)$ is out of the feasible set, then we need to project $(w^0,s^0)$ onto the boundary of the halfspace.
Let $ L=\|\alpha\|_{B_t}^2+\delta\beta^2+\lambda(a^T\alpha-\beta+c-s^0)
$, take the derivative with respect to  $\alpha$ , $\beta$ and   $\lambda$ , make them equal to zero, we get 
\begin{equation}
\begin{cases}
   \frac{\partial L}{\partial\alpha}= 2 B_t\alpha+\lambda \alpha=0\\
   \frac{\partial L}{\partial\beta}= 2\delta\beta-\lambda =0\\
   \frac{\partial L}{\partial\lambda}= a^T\alpha-\beta+c-s^0=0.
\end{cases}
\end{equation}   
To solve these problems we have:
\begin{equation}
 \lambda =\frac{2(c-s^0)}{\frac{1}{\delta}+\|a\|_{B_t^{-1}
}}   , \;\alpha=-\frac{\lambda}{2}B_t^{-1}a,\;\beta=\frac{\lambda}{2\delta}.
\end{equation}
By plugging in and enumerating all possible cases, we get the closed solution \eqref{eq:lemma4}.

\subsection{Proof of Lemma 3 (PSPSL2)}
The slack variables in the objective function of \eqref{eq:pspsl2} can be re-written as 
\begin{equation}
    \mu(s-s_t)^2+\lambda s^2 =\frac{1}{\hat\lambda}(s-\mu\hat\lambda s_t)^2+constant\; w.r.t. \;s ,
\nonumber
\end{equation}
where $\hat\lambda=\frac{1}{\mu+\lambda}$. After dropping constants, solving \eqref{eq:pspsl2} is equivalent to 

\begin{equation}
    w_{t+1},s_{t+1}={\arg\min}_{w\in\mathbb{R}^d,s\in \mathbb{R}}
    \|w-w_t\|_{B_t}^2+
    \frac{1}{\hat\lambda}(s-\mu\hat\lambda s_t)^2  \nonumber
\end{equation}
\begin{equation}
   f_i(w_t)+\left\langle \nabla f_i(w_t),w-w_t\right\rangle\leq s.
\end{equation}
By Lemma \eqref{lemma:lemma5} with $a\leftarrow \nabla f_i(w_t) $,  $c\leftarrow f_i(w_t) $, $s^0\leftarrow \mu\hat\lambda s_t $ and $\delta\leftarrow \frac{1}{\hat\lambda}$, we have the solution given by
Lemma 3.

\end{document}